\documentclass[letterpaper, 10 pt, conference]{ieeeconf}  

\IEEEoverridecommandlockouts                              

\usepackage{times}
\usepackage{epsfig}
\usepackage{graphicx}
\usepackage{amssymb} 
\usepackage{amsmath}
\usepackage[linesnumbered,ruled]{algorithm2e}
\usepackage{array}
\usepackage{rotating}
\usepackage{multirow}
\usepackage{csquotes}
\usepackage{color}
\newcolumntype{P}[1]{>{\centering\arraybackslash}p{#1}}

\newcommand{\thickhline}{%
    \noalign {\ifnum 0=`}\fi \hrule height 1pt
    \futurelet \reserved@a \@xhline
}

\title{\LARGE \bf
Robust and efficient post-processing for video object detection\\
\thanks{This research has been funded by FEDER/Ministerio de Ciencia, Innovación y Universidades/Agencia Estatal de Investigación RTC-2017-6421-7 
and PGC2018-098817-A-I00, DGA T45 17R/FSE and the Office of Naval Research Global project ONRG-NICOP-N62909-19-1-2027.
}}

\author{Alberto Sabater$^{1}$ \hspace{0.5cm} Luis Montesano$^{1,2}$ \hspace{0.5cm}   Ana C.~Murillo$^{1}$
\thanks{$^{1}$ A. Sabater, L. Montesano and A.C. Murillo are with 
DIIS - I3A, Universidad de Zaragoza, Spain. {\tt\small \{asabater, acm\}@unizar.es}
}
\thanks{$^{2}$  L. Montesano is also with Bitbrain Technologies, Zaragoza, Spain. {\tt\small \{luis.montesano\}@bitbrain.com}
}
}

\begin{document}

\maketitle
\thispagestyle{empty}
\pagestyle{empty}

\begin{abstract}

Object recognition in video is an important task for plenty of applications, including autonomous driving perception, surveillance tasks, wearable devices or IoT networks. Object recognition using video data is more challenging than using still images due to blur, occlusions or rare object poses. Specific video detectors with high computational cost or standard image detectors together with a fast post-processing algorithm achieve the current state-of-the-art. This work introduces a novel post-processing pipeline that overcomes some of the limitations of previous post-processing methods 
by introducing a learning-based similarity evaluation between detections across frames. 
Our method improves the results of state-of-the-art specific video detectors, specially regarding fast moving objects, and presents low resource requirements. And applied to efficient still image detectors, such as YOLO, provides comparable results to much more computationally intensive detectors. 

\end{abstract}

\section{Introduction}










Many application fields such as robotics, surveillance or wearable devices require object detection over their embedded camera video streams and efficient algorithms to process them. 
Deep learning based detection approaches, such as the well known YOLO~\cite{redmon2016you} or MaskRCNN~\cite{he2017mask}, have boosted image object detection performance in the recent years. 
However, there is still a large gap between object detection performance in video and images, mainly because video data is more challenging due to numerous artifacts and difficulties such as blur, occlusions or rare object poses.
%

\begin{figure}[!tb]
\centering
\includegraphics[width=.95\linewidth]{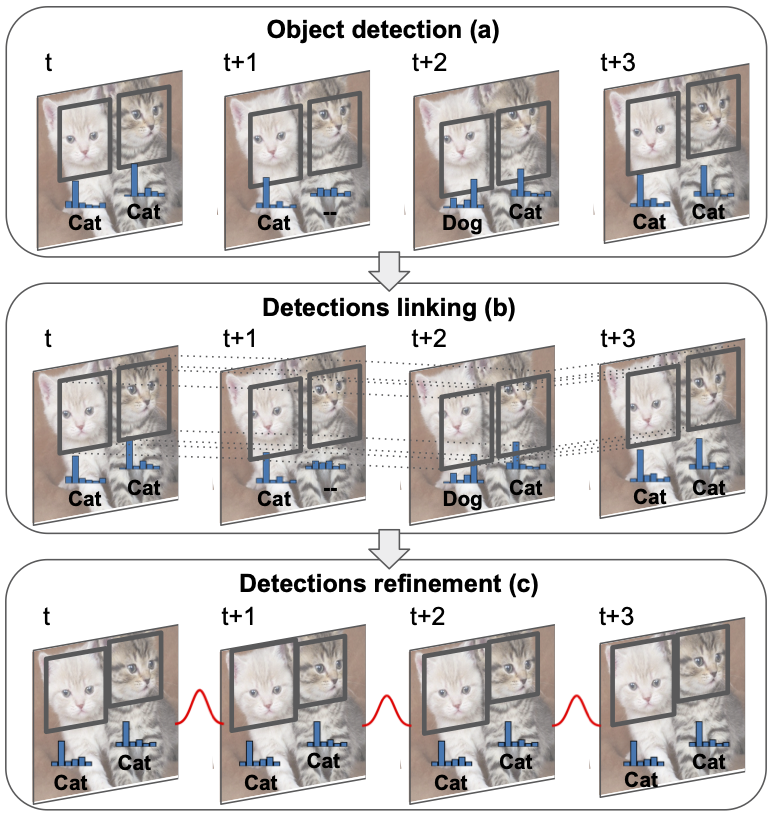}
\caption{Proposed post-processing pipeline to improve video object detection. 
A standard image detector predicts object instances in a sequence of video frames
(a). Our approach  links object instances across frames based on the learned similarity evaluation (b) and uses contextual information to refine the detections (c), both object classification and location.}
\label{fig:intro}
\end{figure}

Two main strategies have been explored to improve object detection in videos. On one hand, there are detection models specifically designed to work over video streams~\cite{Zhu_2017_ICCV, wu2019selsa}. They typically implement feature aggregation from nearby frames and achieve higher accuracy than still image detectors, but they are often slow and require heavy computation.  This makes them not well suited for applications on low-resource environments, like wearable devices or for applications where near real-time computing is a requirement, such as robotics or monitoring video analysis. 
On the other hand, post-processing methods such as Seq-NMS~\cite{SEQ-NMS} and Seq-Bbox-Matching~\cite{SEQ-BBOX:VISIGRAPP} have been proposed to process the outputs of an image object detector evaluated on the video frames to improve the performance. They are mostly based on linking the predicted objects across frames and using these links to refine the detection results. This strategy is typically much faster than specific video object detection methods.


The key for post-processing methods is the way they relate detected objects among consecutive frames. 
This linking is usually based on hand-made heuristics, a common one uses the Intersection over Union (IoU) between object detections. 
This approach has several limitations. 
First, the base detector does not always predict reliable bounding box coordinates. Second, IoU values strongly depend on displacements due to camera or object motion. For instance, fast moving objects may not present enough overlap to get a reliable linking. Note that the same effect occurs when the frame rate drops (e.g., due to computational constraints). Finally, the presence of multiple objects simultaneously in the scene also makes heuristics difficult to design and prone to fail.


This paper presents a novel post processing pipeline for video object detection (see Fig.~\ref{fig:intro}) that can be used in conjunction with any video or image detector. The main novelty relies on the way the similarity is evaluated between object detections to link them across frames. We propose to use a learning-based similarity function that combines different descriptors and is designed to be more robust to varying speeds of object motions. 
Once all possible object instances are linked across frames, a refinement step is run to improve both the classification and location of object detections.

We evaluate our method against state-of-the-art post-processing methods for image detectors on the well known video dataset ImageNet VID~\cite{ILSVRC15}, obtaining better results mainly due to more robust links for fast moving objects. We also show that our method improves the performance of specific video object detectors. Interestingly, the increased robustness to fast moving objects implies we can process frames more sparsely, and then  allows us to use more computationally demanding object detectors even if time constraints are high and not all frames can be processed. Code, learned models and training data are available online\footnote{https://sites.google.com/a/unizar.es/filovi/}.


\section{Related Work}



Object detection in videos often builds on top of image object detection. The latter is a well studied problem with current state-of-the-art methods based on deep learning architectures. Multi-stage detectors~\cite{ren2015faster, R-FCN:NIPS2016} follow R-CNN~\cite{DBLP:journals/corr/GirshickDDM13} and split the prediction process in two stages: candidate selection and candidate classification. Single shot models, on the other hand, use a single Neural Network trained end-to-end to perform object detection in a single step. Many variants exist \cite{liu2016ssd,redmon2016you, Law_2018_ECCV,Zhou_2019_CVPR, zhou2019objects} with different object representations. They are in general faster but perform worse than multi-stage ones. 

There are two types of approaches to extend object detection to video and cope with its specific challenges  (blur, occlusions, rare poses) and to exploit the temporal information and consistency in video data:

\paragraph{Video Detectors}

Video object detectors are designed to exploit the surrounding context of a frame and usually the model propagates or shares object features across frames. 
FGFA~\cite{Zhu_2017_ICCV} aggregates nearby features along the motion path given by an optical flow estimation. D\&T~\cite{Feichtenhofer17DetectTrack} trains a ConvNet end-to-end both for object detection and tracking by using correlation across feature maps and re-scores linked detections. SELSA~\cite{wu2019selsa} extracts object proposals from different frames of a video and aggregates their features depending on their semantic similarities. TSM~\cite{Lin_2019_ICCV} shifts features along the temporal dimension to perform object detection with 2D CNNs. TCD~\cite{DBLP:journals/corr/abs-1811-11167} conditions the output of a single image detector by the tracklets calculated in previous steps.

Video object detectors are usually more computationally expensive than detectors working on still images due to the 
increased network complexity, their need to process more data and often the requirement to calculate additional data such as optical flow. 

\paragraph{Post-processing methods to improve video detection}
Post-processing methods incorporate the temporal context information to the output predictions of either image or video object detectors. 
Applied to per-frame detections, these methods speed-up the inference pipeline with respect to specific video detectors while boosting the final detection performance with respect to their base detector.

Some post-processing methods are based on Kalman Filter variations and use tracking ideas to make the detections more robust or consistent. These are often applied to specific domains like person re-identification, where persons have to be detected and tracked in videos~\cite{DBLP:journals/corr/MilanL0RS16, Chavdarova_2018_CVPR} extracted from cameras usually with a fixed position. Deep SORT~\cite{Wojke2017simple} improves the SORT~\cite{Bewley2016_sort} algorithm, based on Kalman Filters and the Hungarian Algorithm, by adding appearance information to each predicted bounding box.

Other strategies applied in more general settings, are based on bounding box propagation or matching across frames. For example, T-CNN~\cite{TCNN} uses context information to suppress false positives and optical-flow to propagate detections across frames to reduce false negatives. 
Seq-NMS~\cite{SEQ-NMS} matches high overlapping detections across frames within the same clip to make detection results more robust. 
Seq-Bbox-Matching~\cite{SEQ-BBOX:VISIGRAPP} links overlapping detections from continuous pairs of frames to create and re-score object instances and uses them to infer missed detections.
%

Our proposed approach is related to this last group of post-processing techniques. We perform bounding box linking across frames, but instead of hand-made heuristics, a learned classifier is used to distinguish whether two detections belong to the same object instance or not. This model exploits both intermediate features from the base object detector and additional properties of the bounding box.

\section{Proposed Framework}

Our proposed approach for video object detection runs the three modules summarized in Fig.~\ref{fig:intro}, which are detailed next.

\subsection{Object detection and description}
\label{sec:det_and_descr}

Our approach works on top of any initial object detector that can provide object bounding boxes and class confidence score vectors. 
For each video frame $t$, we get a set of object detections, 
and each object detection $o^i_{t}$ 
is described by:
\begin{itemize}
    \item Location and geometry, i.e., its bounding box information: $bb^i_{t} = \{x, y, w, h \}$
    
    \item Semantic information, i.e., the vector of class confidences $cc^i_{t}$ provided by the detector. $cc^i_{t} \epsilon \mathbb{R}^C$, where $C$ is the number of classes within the dataset.

    \item Appearance, i.e., a L2-normalized embedding, $app^i_{t}\epsilon\mathbb{R}^{256}$, representing the appearance of the patch. It is learned as detailed next. 
\end{itemize}

\begin{figure}[!tb]
\centering
\includegraphics[width=0.9\linewidth]{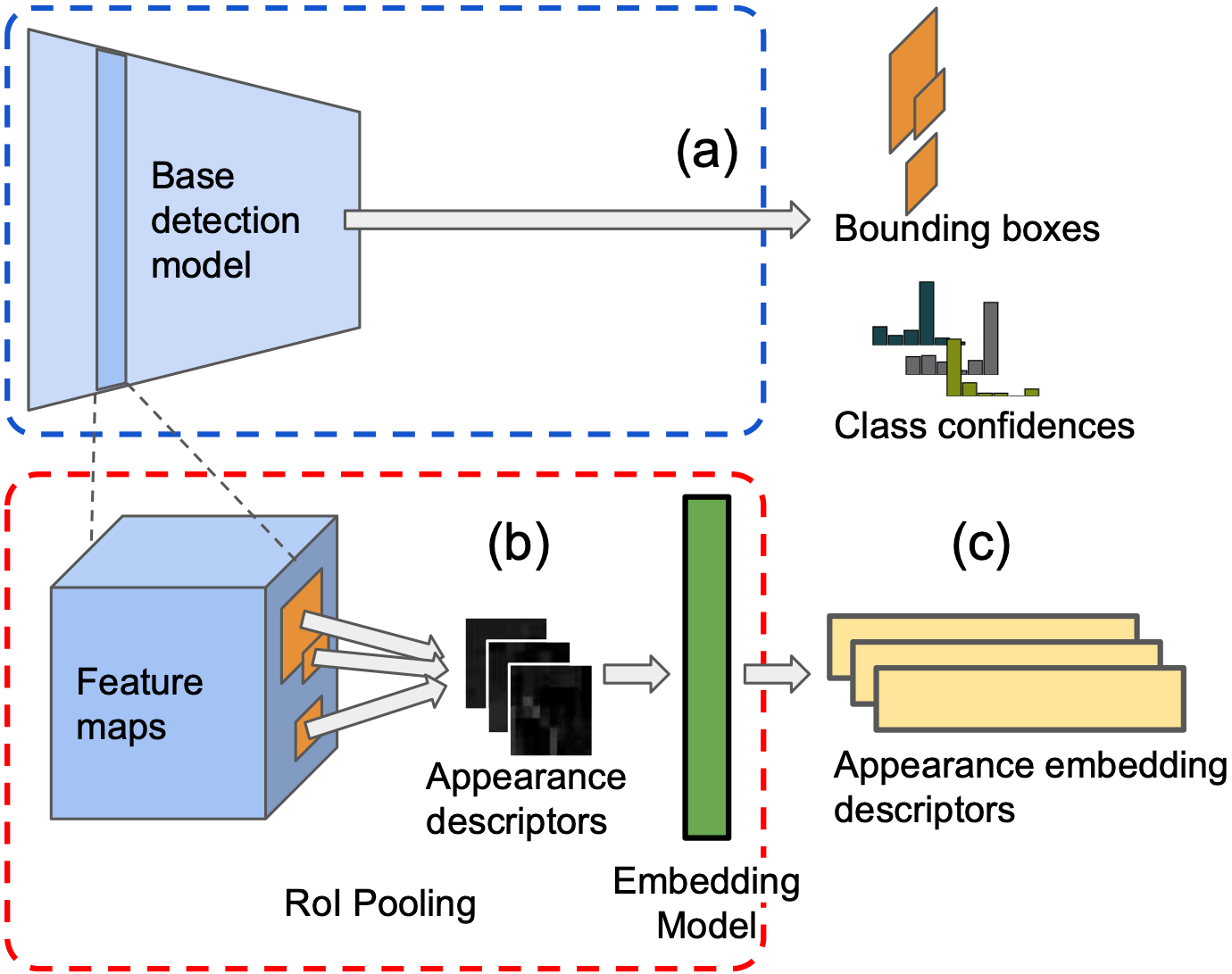}
\caption{
\textbf{Object detection and description}. 
Each detection from the base object detector is represented by the bounding box and vector of class confidences provided by the base model (a) and by an appearance descriptor. 
The appearance descriptor is built from a set of feature maps pooled from the base detector (b) mapped into a lower-dimension embedding vector (c). 
}
\label{fig:descriptors}
\end{figure}

Figure \ref{fig:descriptors} summarizes how these descriptors are obtained. The first two are directly provided by the base object detection model. The appearance descriptor is computed from a set of feature maps generated by an intermediate layer of the base model. 
We propose a simple architecture to learn this appearance embedding as shown in the figure. A RoI Pooling Layer~\cite{ren2015faster} is used to extract 
the feature map outputs that correspond to each of the predicted bounding boxes and scale them to fit a pre-defined shape. 
Since such intermediate descriptors are generally large, we use a single Fully Connected layer to learn a mapping to a lower dimensional embedding and limit the memory and computational resources. 
This embedding model is trained minimizing the triplet loss proposed in~\cite{Schroff_2015_CVPR}:
\begin{equation}
\label{eq:triplet-loss}
\sum^N_i \left [ \left \| f(x^{a}_{i}) - f(x^{p}_{i}) \right\|^{2}_{2} - \left \| ] f(x^{a}_{i}) - f(x^{n}_{i}) \right\|^{2}_{2} + \alpha \right ]_+, 
\end{equation}%
\noindent that compares one Anchor sample \(x^{a}_{i}\) to one Positive sample \(x^{p}_{i}\) and one Negative sample \(x^{n}_{i}\). 
$f(x)\epsilon \mathbb{R} ^d$ embeds a sample $x$ in a $d$ dimensional space and $\alpha$ is the margin enforced between positive and
negative pairs.
The loss minimizes the euclidean distance between the Anchor and the Positive samples, while maximizing the euclidean distance between the Anchor and the Negative sample.  Section~\ref{sec:setup} 
describes the essential step of creating the triplets training set as well as all the other implementation details.

\subsection{Object detection linking}
\label{sec:objlinking}

\begin{figure}[!tb]
\centering
\includegraphics[width=0.9\linewidth]{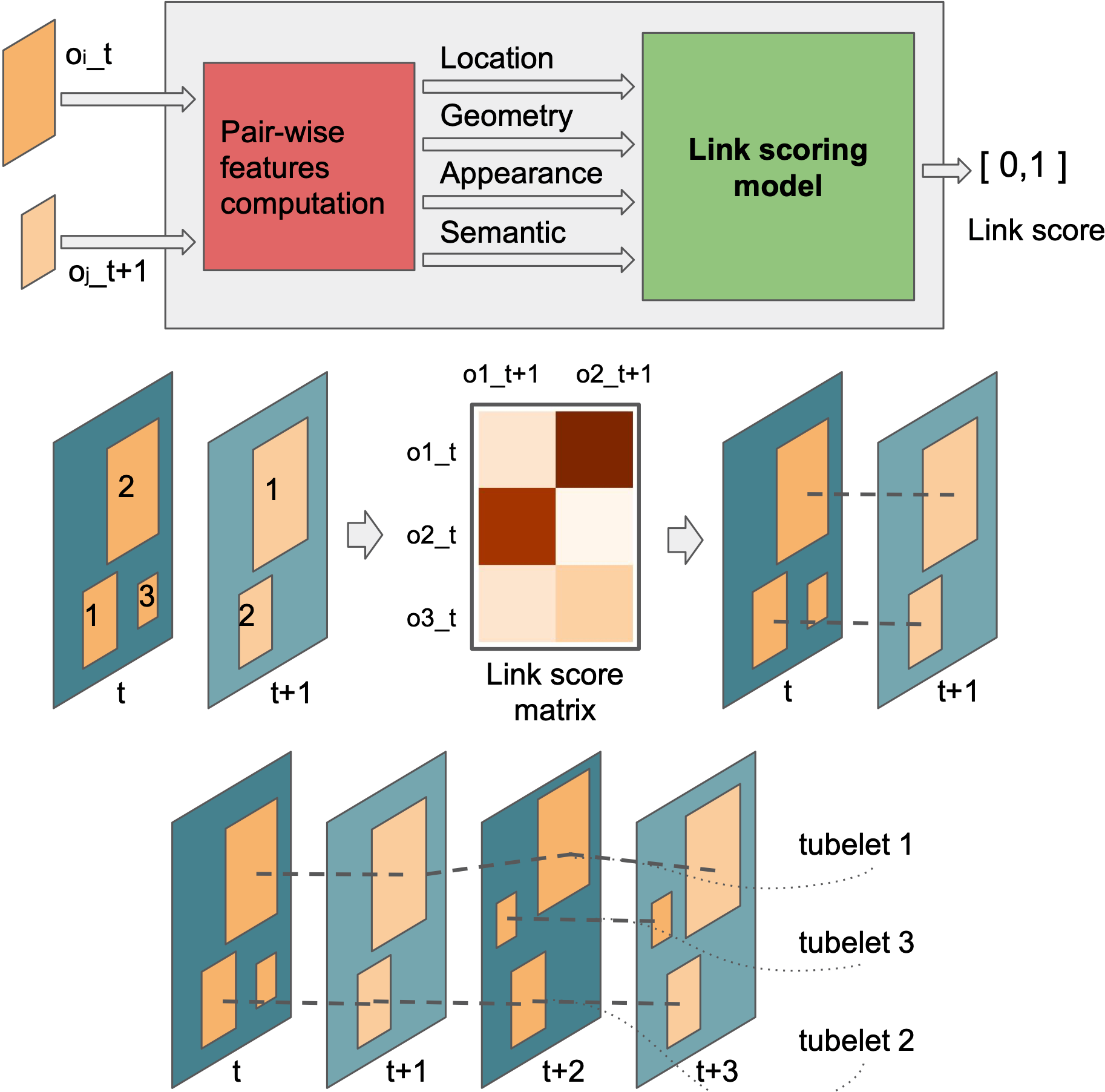}
\caption{
\textbf{Object detection linking}.
For all possible pairs of detections from consecutive frames ($t$ and $t+1$), we build a set of features based on their location, geometry, appearance and semantics. These features are used to predict a linking (similarity) score. Links are established between consecutive frames and tubelets are composed of links as long as possible.
}
\label{fig:linking}
\end{figure}

Our second module links detections into tracks by building a set of tubelets, i.e. sets of corresponding detections along the video, in a similar manner to~\cite{SEQ-BBOX:VISIGRAPP}. This linking, summarized in Fig.~\ref{fig:linking}, is a sequential process. 
We start building tubelets from the object detections between the first pair of frames and extend them as long as corresponding objects are still found in the next following frames. New tubelets can be initialized at any frame with those detections not included in existing tubelets.

To link detections between two consecutive frames, we propose a similarity function based on the following pair-wise features computed for each possible link of detections (\(o^{i}_{t}\) and \(o^{j}_{t+1}\)):%
\begin{eqnarray*}
    f_{loc}     & = & \{IoU, d_{centers} \}, \\
    f_{geo}     & = &\{ ratio_w, ratio_h \}, \\
    f_{app} & = & d_{app}, \\
    f_{sem} & = & f^a_{sem} \cdot f^b_{sem}, 
\end{eqnarray*}

\noindent where $IoU$ is the Intersection over Union of both detections, $d_{centers}$ is the relative euclidean distance between the two bounding box centers, 
$ratio_w$ and $ratio_h$ are the width and height ratio between the two bounding boxes, $d_{app}$ is the euclidean distance between the appearance embeddings and $f_{sem}$ is the dot product of class confidences vectors ($cc_{t}$ and $cc_{t+1}$). 
The actual link score ($LS$) between two detections is computed as follows: 
\begin{equation}
\label{eq:distance-final}
LS(o^{i}_{t}, o^{j}_{t+1}) = f_{sem} \; \textbf{X}(f_{loc}, f_{geo}, f_{app}),
\end{equation}
\noindent where $\textbf{X}$ is a logistic regression trained to distinguish whether two detections, given their pair-wise features, belong to the same object instance (high linking score) or not (low linking score). More details on how this regression is trained based on triplets information can be found in Section~\ref{sec:triplet-dataset}.

A score matrix is created with the link scores of every pair of detections between two frames. We use a greedy approach to match them and extend the tubelet, i.e., the highest score is picked to be a link, and the corresponding row and column are suppressed from the matrix, repeating the process until no more links can be set. 
Although we tried more complex methods to solve the assignment problem, such as the Hungarian method, 
we did not find any improvements to the algorithm described above.
For each pair of frames, a new tubelet is created if an assigned pair did not belong to an already existing tubelet. 

Since object detectors do not always make accurate predictions, some instance linkings are prone to generate false positives when they do not belong to any real object. When this happens, our algorithm outputs a low linking score. Linking candidates that present a score under the established threshold are filtered out.

\subsection{Refinement: re-scoring and re-coordinating}

The final step of our method uses the linkings of each tubelet to improve its object classification and location:

\subsubsection{Re-scoring}
This step simply averages all class confidence vectors from each tubelet and assigns this average to all detections within the tubelet. This process is able to correct mislabeled detections in a subset of frames or disambiguate those with low confidence.

\subsubsection{Bounding box coordinates}
Object detectors predict highly accurate bounding box coordinates on still images or on low-motion frames, but this regression tends to be less accurate on the objects that present rare poses, defocus or fast motions. 
%
%
We treat each coordinate of a linked object over time as a noisy time series. 
Note that noisy bounding box detections cannot properly fit the real object along time.
We use smoothing to remove or alleviate this noise. In particular, we convolve a one-dimensional Gaussian filter along each time series. The smoothed series are then used as the set of coordinates of the object in the tubelet.

\section{Experiments}




This section describes our post-processing evaluation on different detectors and data conditions.

\subsection{Experimental setup}
\label{sec:setup}

\subsubsection{Dataset}
\label{sec:datasets}
The main dataset used is ImageNet VID~\cite{ILSVRC15}, so far the largest densely annotated dataset for video object detection. It consists of 3862 and 555 training and validation snippets densely labeled with multiple bounding boxes belonging to 30 different object classes. These classes are a subset of the 200 classes from ImageNet DET dataset~\cite{ILSVRC15}. ImageNet VID data includes a wide variety of conditions, such as different object movement types, blur, defocus or occlusions. There is an average of 1.59 objects per frame, each one with ground truth annotations of their bounding box and track id, both in train and validation sets.

\subsubsection{Evaluation metrics}
Mean Average Precision (mAP) is computed to evaluate video object detection performance in ImageNet VID, as established in the original paper, and to compare our approach with other methods. To provide deeper insight, we also report the mAP according to three groups of motion (slow, medium and fast) as defined in~\cite{Zhu_2017_ICCV}.

\subsubsection{Triplet dataset used for training our models}
\label{sec:triplet-dataset}

Both the embedding model (Sec. \ref{sec:det_and_descr}) and the link scoring model (Sec. \ref{sec:objlinking}) are trained with the same dataset organized in \textit{triplets}.
Each triplet is composed of one Anchor (sample point), one Positive (same object instance) and one Negative (different object instance) bounding box, 
all of them according to ImageNet VID data groundtruth. Note that for the link scoring model each triplet provides a positive and a negative training example. 

We compiled a set of 50000 and 8000 triplet samples, for training and validation respectively, built as follows. For each triplet sample, we randomly sample a track id, i.e., an object track. The Anchor is obtained from a random frame of the track.  The Positive example is taken from a frame sampled from $\pm$ 25 frames away of the Anchor, i.e. within a one second window. Note that most videos in the dataset are recorded with a frame rate between 20 and 30 frames per second. 
%
The Negative sample is simply an object with a different track id. It can be randomly obtained from either the same snippet as the Anchor or from another video. Negative examples may include objects from the same class but different instance. 

\subsubsection{Base detection model} 
\label{sec:base_det_model}


Although our approach works with any object detector on video data, for comparison purposes we have defined a single baseline, YOLOv3~\cite{redmon2018yolov3}, as our per-frame base object detector.
YOLO is a well known and broadly used single-shot detection model that uses a Fully-Convolutional Neural Network architecture to get predictions at different scales, achieving great time-accuracy trade-off.

This base model has been trained with data both from ImageNet DET and ImageNet VID using the same data split as FGFA~\cite{Zhu_2017_ICCV}. We trained the model with data augmentation techniques:  multi-scale input, horizontal flip, cropping, shifting, jitter and saturation. For the inference phase we fix the input image shape to 512x512 pixels and we replace the usual non-maximum-suppression (NMS) implementation, that works at a class detection level. Instead, we apply it at a bounding box level taking the maximum class confidence score for each detection in a frame as one of the inputs of NMS. Our NMS module finally outputs a set of bounding boxes for each frame along with a class confidence vector for each bounding box.
Finally, we filter out those detections whose maximum class confidence is under a threshold of 0.005. The predictions obtained in this inference phase are the same data that we use in all the tests we perform within our ablation study and comparison with other post-processing techniques. With this configuration we are able to obtain predictions at 23 fps with a single NVIDIA GeForce RTX 2080Ti GPU.


\subsubsection{Appearance embedding model configuration}
The feature maps used for the appearance embedding correspond to the output of the Convolutional Layers Block from the base detection model, that downsamples the input image by a factor of 16. Our RoI Pooling extracts the feature patches from these maps and scales them to a shape of 5x5x256. 
These values are set to minimize the loss of information on the scaling phase while allowing to predict fast appearance embeddings. The Fully Convolutional layer that learns the final embedding contains 256 neurons and outputs $L2$-normalized embeddings of 256 values.


\subsection{Performance \& analysis of other post-processing methods}

\begin{table*}[!tb]
\caption{
Results of different post-processing approaches for object detection in ImageNetVID validation set.}
\centering
\begin{tabular}{l|c|c|c|c|c|}
\cline{2-6}
& \textbf{mAP} 
& \multicolumn{1}{p{1.8cm}|}{\textbf{mAP - slow motion objects}} 
& \multicolumn{1}{p{1.9cm}|}{\textbf{mAP - medium motion objects}} 
& \multicolumn{1}{p{1.8cm}|}{\textbf{mAP - fast motion objects}} 
& \multicolumn{1}{p{4.8cm}|}{\textbf{avg. processing time (ms) per frame (detection + post-processing)}}\\
\hline
\multicolumn{1}{|l|}{\textbf{Only base detector$^{+}$}} & 68.59 & 76.79 & 66.45 & 45.79 & 44 \\ \hline
\multicolumn{1}{|l|}{\textbf{Ours }} & \textbf{75.06} & \textbf{82.54} & \textbf{74.29} & \textbf{56.58} & 46.58 (44 + 2.58) \\ 
\hline
\multicolumn{1}{|l|}{\textbf{Seq-NMS~\cite{SEQ-NMS}}} & 71.51 & 79.99 & 70.01 & 50.95 & 54.41 (44 + 10.41) \\ \hline
\multicolumn{1}{|l|}{\textbf{Seq-Bbox-Matching~\cite{SEQ-BBOX:VISIGRAPP}}} & 74.19 & 81.13 & 73.22 & 54.39 & 44.40 (44 + 0.40) \\ \hline
\multicolumn{6}{@{}p{16cm}}{\footnotesize{$^{+}$All other methods are run as post-processing of this base detector.}}\\
\end{tabular}
\label{tab:ablation}
\end{table*}

\begin{figure*}[!tb]
\centering
\includegraphics[width=1.\linewidth]{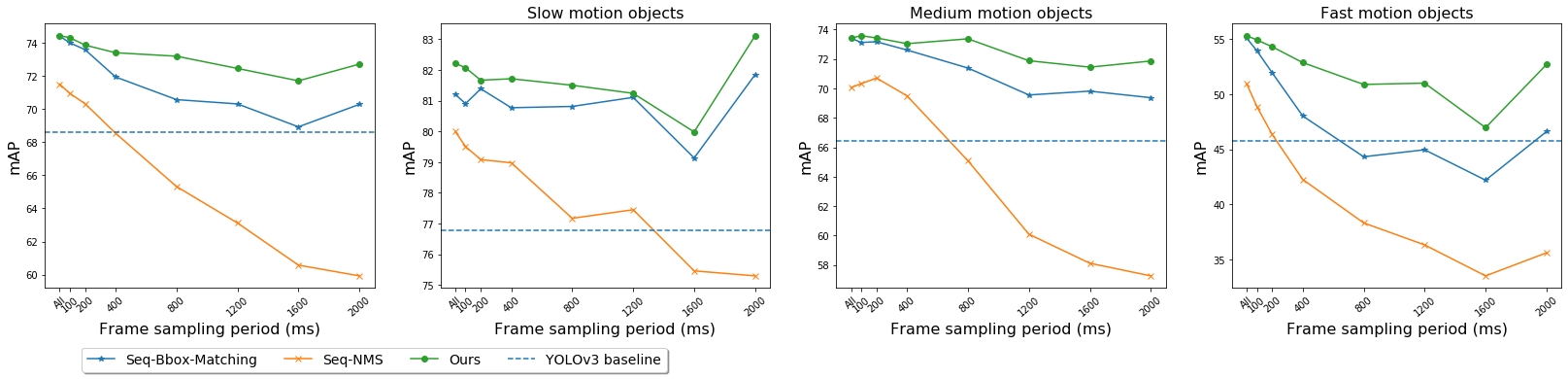}
\caption{Performance of different post-processing methods evaluated with different frame sampling period. (a) Shows mAP on all object instances. Following~\cite{Zhu_2017_ICCV} (b), (c), and (d) shows the mAP of the objects according to their motion, slow, medium of fast. Dotted line shows the mAP baseline for each motion. All post-processing methods have been applied to our YOLOv3 baseline predictions. 
Seq-NMS has been calculated with the code released by~\cite{Zhu_2017_ICCV}.
Seq-Bbox-Matching has been calculated with code replicated from the paper, since the original one is not available.}
\label{fig:sparse}
\end{figure*}


 Table~\ref{tab:ablation} shows object detection performance of our post-processing approach compared to two other well known approaches on the ImageNetVID dataset. We evaluate the improvements of the post-processing methods with respect to the predictions obtained from a common base model (YOLOv3). We measure the mAP for all test videos and detail specific results for slow, medium and fast moving objects.  The average processing time per frame is also shown. Each row corresponds to one of the following approaches:

\begin{itemize}

\item{\textbf{Only base detector}.} Corresponds to the execution of the base model detector, YOLOv3, with the configuration described in Sec.~\ref{sec:base_det_model} with no post-processing.
\item{\textbf{Ours.}} Corresponds to YOLOv3 detections post-processed with our proposed method configured with the following parameters: linking threshold of $0.7$ to suppress low-scoring detection linkings, and standard deviation of the Gaussian filter in the re-coordinating module set to $0.6$.
\item{\textbf{Seq-Bbox-Matching~\cite{SEQ-BBOX:VISIGRAPP}.}} Corresponds to YOLOv3 detections post-processed with Seq-Bbox-Matching. Since it does not have a public implementation, we have replicated it from the paper (\(\kappa\) value of 12 for their tubelet linking module, as they specify).
\item{\textbf{Seq-NMS~\cite{SEQ-NMS}}.} Corresponds to YOLOv3 detections post-processed with Seq-NMS using the public implementation by FGFA~\cite{Zhu_2017_ICCV}, with the same parameters as in the original publication~\cite{SEQ-NMS} but with our own confidence threshold (defined in Section~\ref{sec:base_det_model}). 

\end{itemize}




\noindent The results show how post-processing is able to improve the baseline detector. Our method achieves better performance than the two state-of-the-art methods when applied to the same base detections as our approach. The computation times of our method are slightly higher than those of Seq-Bbox, but they are still in the order of milliseconds. 
Although our method overperforms all the groups, the difference between the methods is more noticeable for fast-moving objects, likely because both previous methods rely too much on the IoU between object detections in consecutive frames, while our approach has additional descriptors and a learned function that make the linking process more robust to changes in position. Our method improves $6\%, 8\%$ and $11\%$ w.r.t. the baseline method, while the other methods improve around $6-7\%$ for all classes. 

To get more insight about this increased robustness to speed, we simulated a lower processing frame rate. This is 
important in practice when limited computational resources impede the processing at the acquisition frame rate. 
This more challenging scenario 
shows how the approaches work when objects change more drastically both their location and appearance. 
We ran the same object detection task on the ImageNet VID validation set, but simulating a lower processing frame rate. Since the videos have different sampling rates, we fixed the time between frames and picked the closest frame for each video.  
We removed tracks composed by less than 2 frames, since post-processing methods are not useful there. The number of videos dropped from 555 to 522 in the smallest data configuration (2000 ms). 
Note that evaluating different frames and tracks for different sampling rates can lead the mAP to not decrease as expected when the dropped data belongs to objects that are difficult to detect (this effect is more noticeable between the frame sampling periods $1000$ and $2000$).
Since linking scores drop their value for long time-distant detections due to their lower similarities, for this test we set a linking threshold of 0.05,
more suitable to suppress spurious detections and being able to get long-term linkings. We also remove the tubelet linking module of Seq-Bbox-Matching and set the parameters of the method according to the paper recommendations for this type of experiment.

\begin{table*}[ht!]
\caption{Comparison with Video Object Detectors on ImageNetVID validation set.}
\centering
\begin{tabular}{|l||lp{1.2cm}||p{0.6cm}|p{0.6cm}|p{0.9cm}|p{0.6cm}|p{4.3cm}|}
\hline
  & 
 \textbf{base object detector} (backbone) & \textbf{mAP backbone} & \textbf{mAP ALL} & \textbf{mAP slow} & \textbf{mAP medium} & \textbf{mAP fast} & \textbf{avg. processing time (ms) per frame (detection + post-processing)} \\ \hline
\textbf{FGFA~\cite{Zhu_2017_ICCV}*} & R-FCN (ResNet101)& 74.1 & 77.1 & 85.9 & 75.7 & 56.1 & 128  \\ \hline
\textbf{TCD~\cite{DBLP:journals/corr/abs-1811-11167}} & Faster R-CNN (ResNet101) & 74.6 & 83.5 &  &  &  & \\ \hline
\textbf{SELSA~\cite{wu2019selsa}*} & Faster R-CNN (ResNet101) & 73.62 & 82.69 & 88.02 & 81.34 & 67.17 & 458  \\ \hline
\hline
\textbf{YOLOv3 + Ours}  & YOLOv3 (Darknet-53) & 68.59 & 75.06 & 82.54 & 74.29 & 56.58 & 46.58 (44 + 2.58)  \\ \hline
\textbf{FGFA + Ours} & R-FCN (ResNet101) & 74.1 & 80.09 & 87.42 & 79.1 & 61.38 & 149 (128 + 21) \\ \hline
\textbf{SELSA + Ours}  & Faster R-CNN (ResNet-101) & 73.62 & 84.21 & 88.72 & 83.32 & 71.09 & 466.6 (458 + 8.6)  \\ \hline
\multicolumn{8}{p{15cm}}{\footnotesize{$^{*}$Inference times have been calculated by running their official code on a single NVIDIA GEOFORCE RTX 2080Ti GPU}}\\
\multicolumn{8}{p{15cm}}{\footnotesize{All models make use of both Imagenet DET and VID data for training (or pretraining).}}
\end{tabular}
\label{tab:video-comparison}
\end{table*}

Figure~\ref{fig:sparse} shows the mAP for all test videos, including separate plots for different object motion speeds, for varying values of frame sampling periods. We can observe similar effects to before. 
Our method gets comparable results to the others when working with continuous frames, but it manages to get more robust results when frames are processed more sparsely. Interestingly, our method is always able to improve over the baseline detector performance, while the other methods cannot. 


\subsection{Post-processing of video object detection methods}

Table~\ref{tab:video-comparison} compares the results of state-of-the-art specific Video Object Detectors with our post-processing method applied to an Image Object Detector (YOLOv3~\cite{redmon2018yolov3}) and to two Video Object Detectors (SELSA~\cite{wu2019selsa} and FGFA~\cite{Zhu_2017_ICCV}). The results with YOLOv3 are the same as in the previous Table~\ref{tab:ablation} and are included to ease comparisons. TCD~\cite{DBLP:journals/corr/abs-1811-11167} results can not be fully analyzed, since there is no code available up to our knowledge. We show their overall result as reference, as reported in the original paper. SELSA and FGFA results have been obtained using their official code implementation and parameters. They use a ResNet-101 as backbone and train with data augmentation and a mix of ImageNet VID and DET. Since extracting low-level features from a Video Object Detector architecture is not trivial, appearance features have not been used in SELSA and FGFA tests.

Table~\ref{tab:video-comparison} shows that post-processing the YOLO single image detections cannot beat the best more complex detectors that work over multiple frames. 
But it is interesting to note that it does achieve slightly better performance than the corresponding much more complex ResNet based models when performing per-frame detections. 
However, when applied to state-of-the-art video object detectors, our approach is still able to boost the performance of SELSA and FGFA between 2 and $3\%$. This gain is mainly due to medium and fast object movements, where we get a 5\% gain. This is not always the case for other post-processing methods that failed to improve video detectors~\cite{wu2019selsa}.

Finally, it is worth stressing again that applying our post-processing approach is always a computationally cheap operation when compared to the detection time.  Post-processing time depends on the number of detections per frame, 
but even with many detections 
it is still in the order of tens of milliseconds. 
Our results show that simple and efficient post-processing methods can boost performance of many state-of-the-art detectors with minimum computational requirements and, when time constraints are important, can be a useful approach to trade-off time performance and computational resources.

\subsection{Qualitative results on EPIC-KITCHEN Dataset}

\begin{figure*}[!htb]
    \centering
    \includegraphics[width=.245\linewidth]{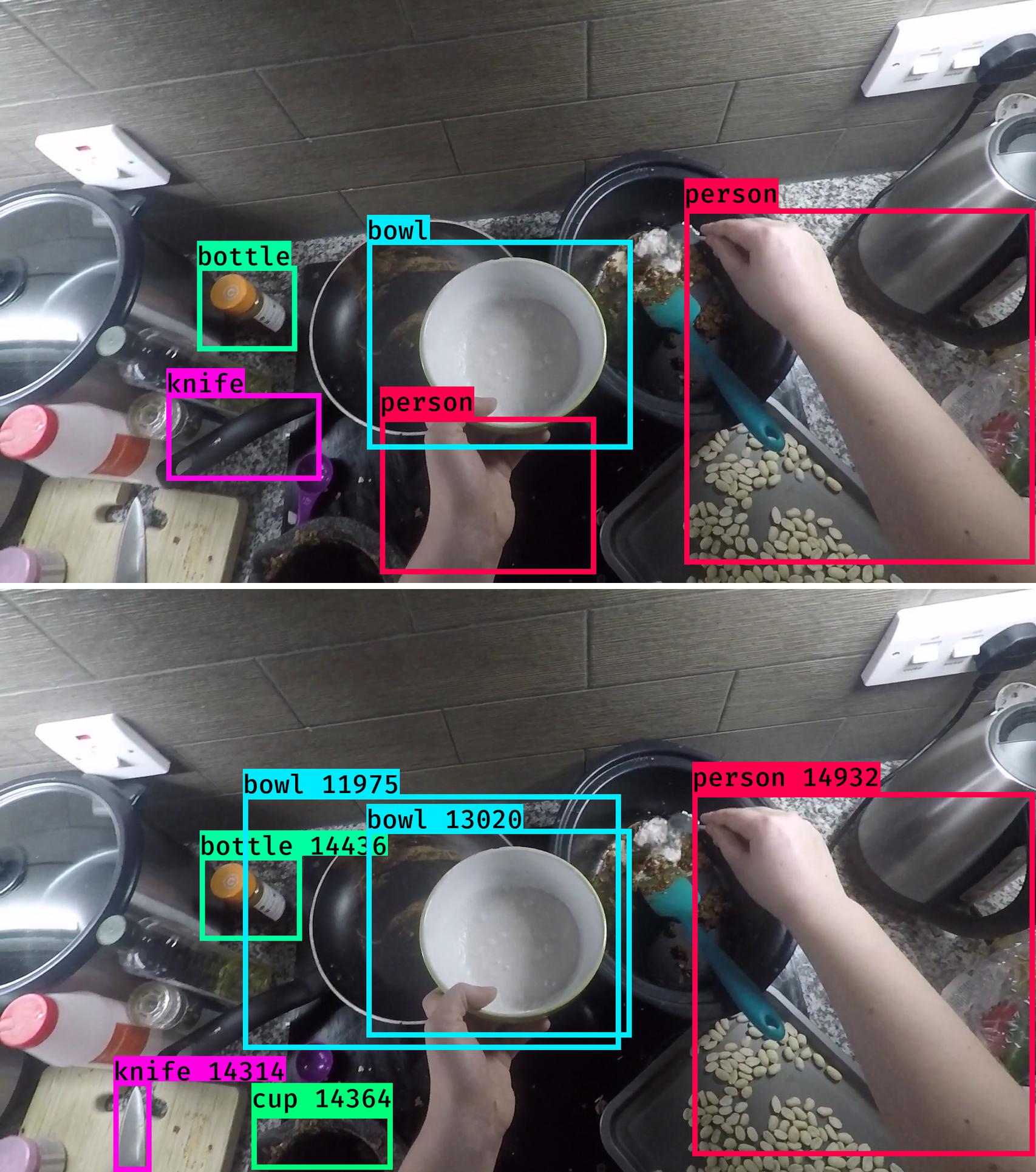}
    \includegraphics[width=.245\linewidth]{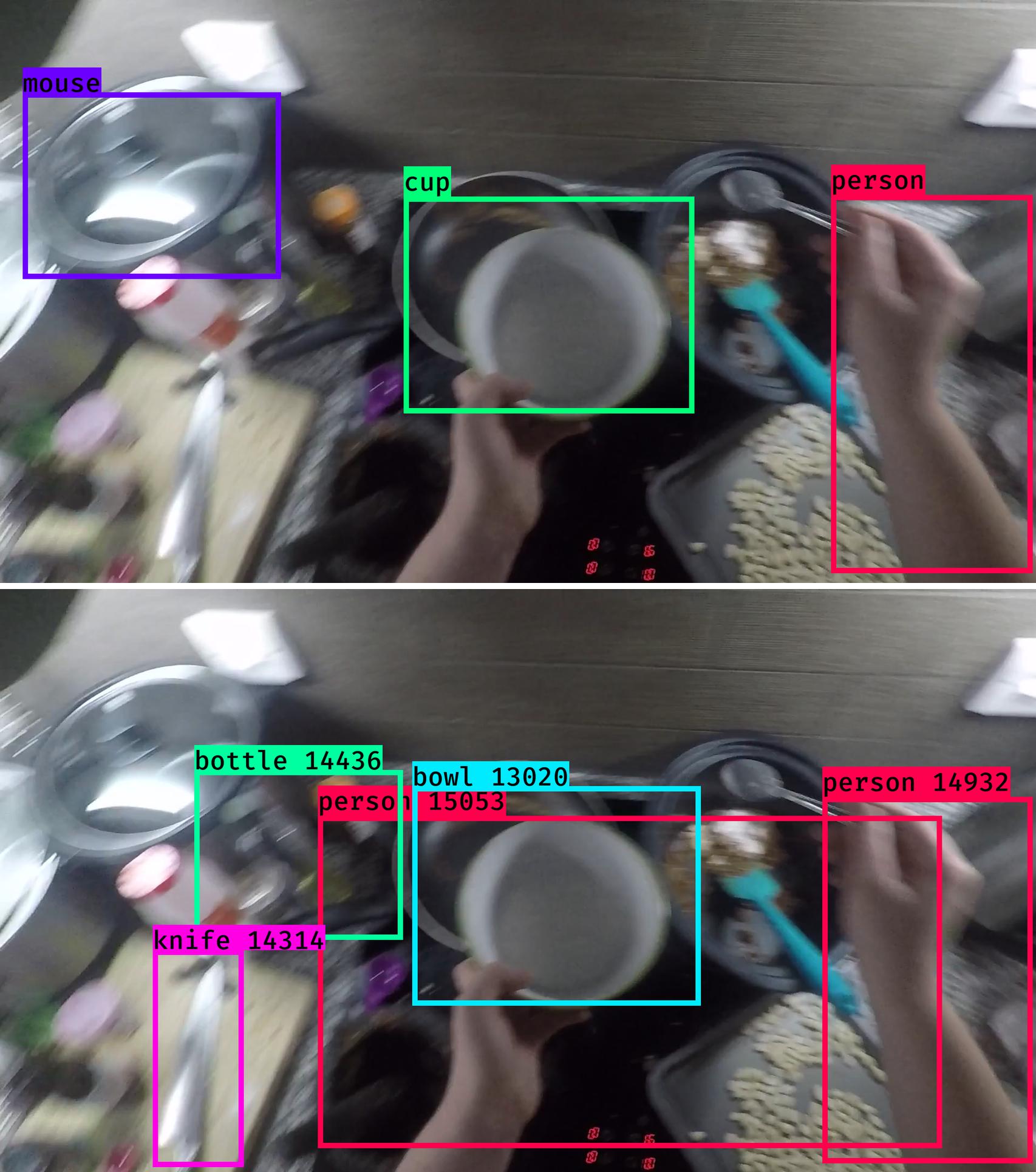}
    \includegraphics[width=.245\linewidth]{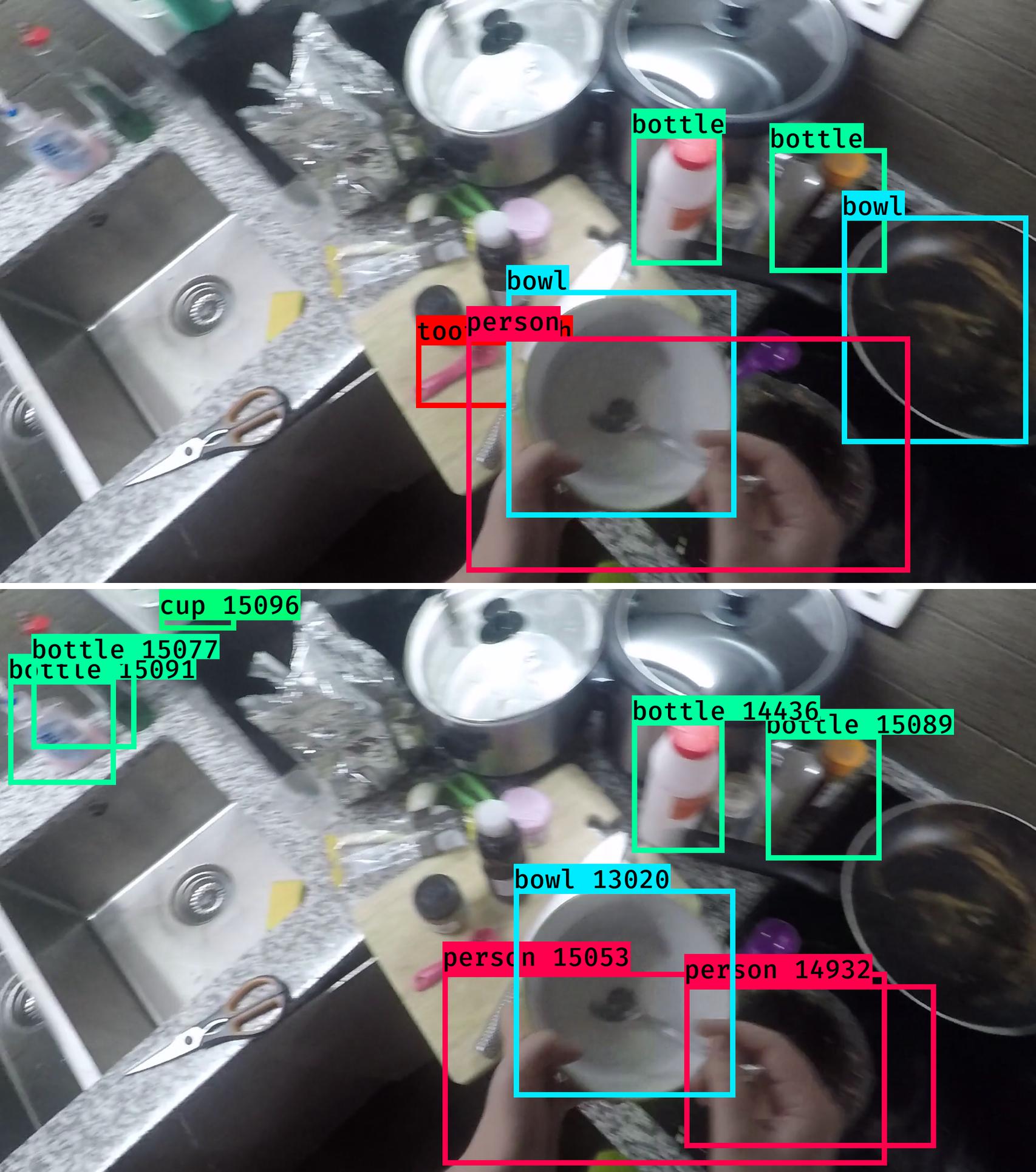}
    \includegraphics[width=.245\linewidth]{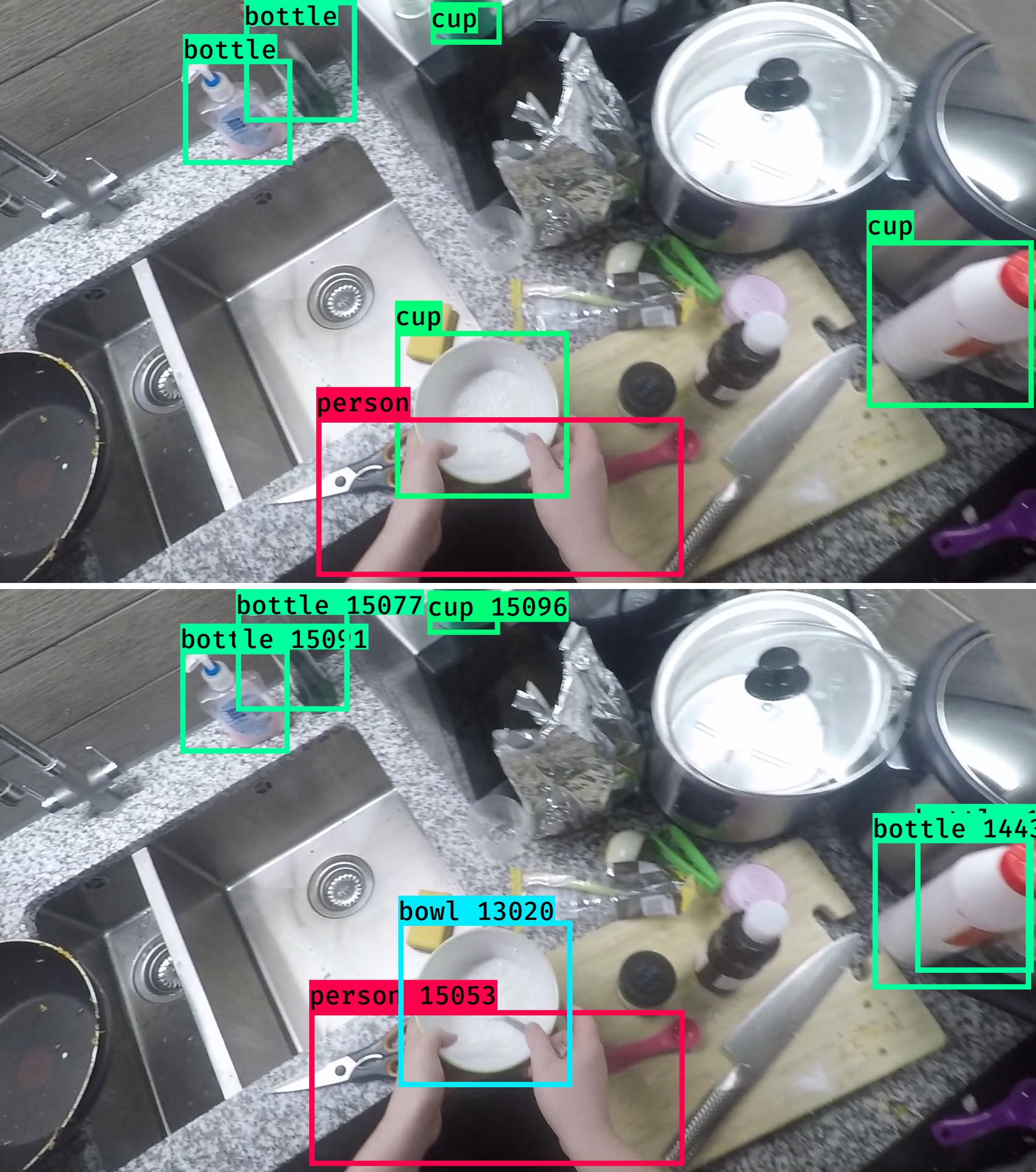}

    \caption{
    Object detection on EPIC-KITCHENS data using a YOLOv3 model (top row) and improvements obtained applying our post-processing (bottom row). Predicted bounding boxes are shown along with their track\_id when our post-processing is applied. To see more varied results these four frames are not consecutive ($18812$, $18832$, $18868$ and $18908$), watch the supplementary video for results on the whole  sequence\protect\footnotemark. 
    Low-scoring detections have been removed for a better visualization
    }
    \label{fig:kitchen_preds}
\end{figure*}

This experiment tests our proposed post-processing pipeline in a more challenging scenario. We perform a qualitative evaluation on the EPIC-KITCHENS dataset~\cite{damen2018scaling}. It is an egocentric video dataset where different users perform daily activities in a kitchen. 
Multiple objects of different categories appear in the scenes with uncommon perspectives and movements compared to traditional video data. Groundtruth includes user action annotations, multi-language narrations and \textit{active} object detections (note this means not all objects are annotated in the videos). 

In order to prove the generalization ability of our approach, we apply it to predictions on these egocentric videos. 
Predictions are obtained with a YOLOv3 model trained only with the COCO dataset~\cite{COCO:ECCV}, just still images from 80 different categories involved.


Figure~\ref{fig:kitchen_preds} shows a scene (frames $18812$, $18832$, $18868$ and $18908$ from the video "P04\_24"; note that KITCHENS videos are recorded at 60 fps) that involves a fast camera motion, where objects drastically change their location and get out of focus. The YOLO predictions (first row) suffer from many misdetections and misclassifications due to the mentioned video artifacts. The application of our approach (second row) manages to correct many of them. Partially occluded objects are detected (tracks $14314$ and $14364$), out of focus objects are correctly detected (track $14436$) and classified (track $13020$) and false positives are suppressed (observe detected knife, mouse and toothbrush from the base predictions). By using neighboring frame information, object coordinates show a smoother evolution along the full video sequence thanks to our re-coordinating module, removing wrong shapes and flickering.



\section{Conclusions}

In this work, we present a novel post-processing method for video object detection. By using a novel set of detection features we study the similarity between frame detections from a learning-based approach as a preliminary step to a prediction refinement. With a light computational overhead, we boost the performance of state-of-the-art video object detectors and applied to efficient still image object detectors we achieve comparable results to more complex models. As demonstrated, our novel post-processing solution allows to overcome the main video challenges such as misdetections and misclassifications due to fast motion objects, occlusions or defocus, and also proves its robustness on environments where lower processing frame rates are a restriction.

\footnotetext{https://youtu.be/QXL2FlyO1jA}


{
\bibliographystyle{IEEEtran}
\bibliography{biblio}
}

\end{document}